\documentclass{IEEEtran}
\usepackage{cite} 
\usepackage{url} 
\usepackage[utf8]{inputenc} 
\usepackage{booktabs} 
\usepackage{graphicx}
\usepackage{authblk}
\usepackage{algorithm}
\usepackage{amsmath}
\usepackage[noend]{algpseudocode}
\usepackage[justification=centering]{caption}
\usepackage{hyperref}
\usepackage{subfigure}

\title{Continual Match Based Training in Pommerman: Technical Report}

\makeatletter
\newcommand{\printfnsymbol}[1]{%
  \textsuperscript{\@fnsymbol{#1}}%
}
\makeatother

\author[1]{Peng Peng\printfnsymbol{1}\thanks{* Equal contribution.}}
\author[2]{Liang Pang\printfnsymbol{1}}
\author[3]{Yufeng Yuan}
\author[1]{Chao Gao}
\affil[1]{Inspir.ai, Beijing, China}
\affil[2]{Institute of Computing Technology, CAS, Beijing, China}
\affil[3]{Beijing Normal University, Beijing, China}
\date{}
\begin{document}
\maketitle

\IEEEpeerreviewmaketitle
\begin{abstract}
Continual learning is the ability of agents to improve their capacities throughout multiple tasks continually. While recent works in the literature of continual learning mostly focused on developing either particular loss functions or specialized structures of neural network explaining the episodic memory or neural plasticity, we study continual learning from the perspective of the training mechanism.
Specifically, we propose a COnitnual Match BAsed Training (COMBAT) framework for training a population of advantage-actor-critic (A2C) agents in Pommerman, a partially observable multi-agent environment with no communication.

Following the COMBAT framework, we trained an agent, namely, \emph{Navocado}, that won the title of the top 1 learning agent in the NeurIPS 2018 Pommerman Competition.
Two critical features of our agent are worth mentioning.
Firstly, our agent did not learn from any demonstrations.
Secondly, our agent is highly reproducible. 
As a technical report, we articulate the design of state space, action space, reward, and most importantly, the COMBAT framework for our Pommerman agent.
We show in the experiments that Pommerman is a perfect environment for studying continual learning, and the agent can improve its performance by continually learning new skills without forgetting the old ones. Finally, the result in the Pommerman Competition verifies the robustness of our agent when competing with various opponents.
\end{abstract}

\section{Introduction}
\label{sec:intro}

Pommerman~\cite{resnick2018pommerman} is a multi-agent environment based on the classic console game Bomberman. Every battle starts on a randomly drawn symmetric 11x11 grid with four agents each initially located in one of the four corners. On every turn, each agent can execute one of the six actions: \emph{Stop, Up, Down, Left, Right and Lay a bomb}. Besides the agents, each cell on the grid can be a passage, a wooden wall, or rigid walls. Both rigid and wooden walls are impassible, whereas wooden walls are destructible with the bomb explosion. After a wooden wall was destroyed, there is a 50\% for it to become a passage and another 50\% for it to reveal one of the hidden power-ups. There are three kinds of power-ups in games: \emph{Extra Ammo, Extra Range and Can Kick} which increases the agent's current ammo by one, increases its bomb blast range by one and enables an agent to kick a bomb, respectively.
There are three variants of the Pommerman environment, i.e., FFA(Free For All), Team and Team Radio. The NeurIPS 2018 Pommerman Competition is about the Team environment, where each participant controls two agents initially located in the corner at the same diagonal as a team, and each agent only observes a 7x7 area centered on its position.

There are several challenges for learning an agent in the Pommerman environment.
First of all, the variance of the game episode length is high. A battle can have as many as several hundred steps when two matched teams play against each other, whereas it may also end shortly with dozens of steps if two teams are badly mismatched.
Secondly, the reward signal obtained at the end of each battle is particularly sparse and delayed. Since the Team environment is an environment mixed of the competitive and cooperative agents, the usual way of learning from such sparse and delayed reward signal is far away from enough.
Thirdly, the features of partially observable and no communication in the Team environment makes it hard for an agent to plan without the full view of the game.
Fourth and most importantly, there are so many skills that an agent can learn in such a complex environment in order to play well against different types of opponents.
Therefore, we regard the Pommerman environment as a good test field of continual learning~\cite{parisi2018continual, thrun1995lifelong, hassabis2017neuroscience}.


In this paper, we propose a COnitnual Match BAsed Training (COMBAT) framework for continual learning in Pommerman.
The idea is inspired by the Population Based Training (PBT)~\cite{jaderberg2017population}, where a population of agents is optimized asynchronously in parallel with a fixed budget of computational resources.
Similar to PBT, COMBAT is a meta-optimization technique, which balances the use of computational resources between exploring new hyperparameters and exploiting the best agent.
However, there are two significant differences between COMBAT and PBT.
On the one hand, unlike PBT, COMBAT does not evaluate each agent's performance in the population independently. 
Specifically, COMBAT uses a round-robin method to generate a schedule of matches, according to which each agent in the population competes with a specified opponent. We update the ELO score~
\cite{balduzzi2018re} to evaluate each agent according to the match results and the ranking list are updated accordingly.
As time evolves, only the agents at the top of the ranking list have an opportunity to learn continually, whereas those at the bottom of the ranking list will be eliminated from the population.
Through such a competition based training mechanism, agents with better hyperparameters stand out, and the population of agents evolved into a more and more competitive community over time.
On the other hand, COMBAT stops exploiting strong agents with the same group of hyperparameters after the performance of these agents was converged for a certain amount of period, but continue to train them by exploring new hyperparameters and reward functions.
The intuition behind is that for any complex problem such as Pommerman, the convergence of a particular partially observable Markov decision making (POMDP)~\cite{hansen2004dynamic, littman2009tutorial} or stochastic game (SG)~
\cite{littman1994markov,myerson2013game} of the problem does not mean the problem has been solved. Instead, by updating hyperparameters and reward functions, it is possible for the agent to retain the hope of reaching the global optimal solution of the problem.

The hyperparameter fine-tuned in our implementation is the discount factor, which is crucial for our agent to keep on improving when the convergence of local optimal is met. 
Since the goal of optimization is the exponentially decayed sum of future rewards, weighted by the exponentially decayed discount factor, the value of the discount factor is closely related to the objective of the optimization algorithm.
More specifically, the discount factor determines how long the future rewards reach its half-life.
For example, when the discount factor is 0.9, the future reward in after $7$ steps is a half-life of its original value; when the discount factor is 0.99, the future reward after $68$ steps is a half-life of its original value.
Empirically, we found that with different discount factors, Pommerman agents are prone to learn different skills. The agent with a relatively smaller discount factor is inclined to learn the reactive skills such as escaping from the bomb explosion, while the agent with a relatively larger discount factor is prone to learn the non-reactive skills such as navigating to pick up a power-up.
Meanwhile, the reward function should also match the use of discount factor. During the period when the agent learned a particular skill, the corresponding reward function should also be highlighted.

The roadmap of this report is as follows.
In Section~\ref{sec:prodef}, we give the problem definition.
Next, we present the details of our agent and the COMBAT framework in Section~\ref{sec:method}.
In Section~\ref{sec:related}, we review related work on continual learning and hyperparameter search.
In Section~\ref{sec:discuss}, we further discuss some interesting results in Pommerman and COMBAT.
Finally, the experimental results are shown in Section~\ref{sec:exp}.

\section{Problem Definition}
\label{sec:prodef}

Even though the Pommerman environment is partially observable, since the vision of a Pommerman agent (7x7) is relatively large compared with the size of the map (11x11) and the focus of this work is not about partial observability,
we consider the problem as a Markov Decision Process (MDP)~\cite{puterman2014markov, sutton2018reinforcement}, a widely used sequential decision-making model.
An MDP is composed of states, actions, rewards, policy, and transitions, and represented by a tuple $\langle S, A, T, R, \pi \rangle$. 

\begin{itemize}
    \item \textbf{States} $S$ is a set of discrete or continuous states
    \item \textbf{Actions} $A$ is a discrete set of actions that an agent can take. The actions available may depend on the state $s$, denoted as $A(s)$.
    \item \textbf{Transition} $T$ is the state transition function $s_{t+1} = T(s_t, a_t)$ which specifies a function which maps a state $s_t$ into a new state $s_{t+1}$ in response to the action selected $a_t$.
    \item \textbf{Reward} $R(s, a)$ is the immediate reward.It gives the immediate reward of taking action $a$ at state $s$. 
    \item \textbf{Policy} $\pi(a|s)$ describes the behaviors of an agent, which is a probability distribution over the possible actions. $\pi$ is usually optimized to decide how to move around in the state space to optimize the long term return.
\end{itemize}

The agent and environment interact at each of a sequence of discrete time steps, $t=0, 1, 2, \cdots$. At each time step $t$ the agent receives some representation of the environment's state, $s_t\in S$, and on that basis selects an action $a_t\in A(s_t)$, where $A(s_t)$ is the set of actions available in state $s_t$. One time step later, in part as a consequence of its action, the agent receives a numerical reward, $r_{t+1}\in R$ and finds itself in a new state $s_{t+1} = T(s_t, a_t)$. 


The agent’s goal is to find a policy that maximizes
the return at every single step. So, we define an objective function $J$ that allows us to score an arbitrary policy parameter. 
\begin{equation}
    \begin{aligned}
        J_\theta(\pi_\theta) & = \int_{\mathcal{S}} \rho^\pi(s)\int_{\mathcal{A}}pi_\theta(a|s)Q^\pi(s,a)
    \end{aligned}
\end{equation}

By using the policy gradient algorithm~\cite{sutton2000policy},
the gradient of the objective function $J$ with respect to parameter $\theta$ can be written as:
\begin{equation}
    \begin{aligned}
        \nabla_\theta J_\theta(\pi_\theta) & = \int_{\mathcal{S}} \rho^\pi(s)\int_{\mathcal{A}}\nabla_\theta\pi_\theta(a|s)Q^\pi(s,a)dads \\
        & = E_{s\sim \rho^\pi, a\sim \pi_\theta} [\nabla_\theta\log\pi_\theta(a|s)Q^\pi(s,a)]
    \end{aligned}
\end{equation}

\section{Methods}
\label{sec:method}

\begin{figure}
  \includegraphics[width=0.45\textwidth]{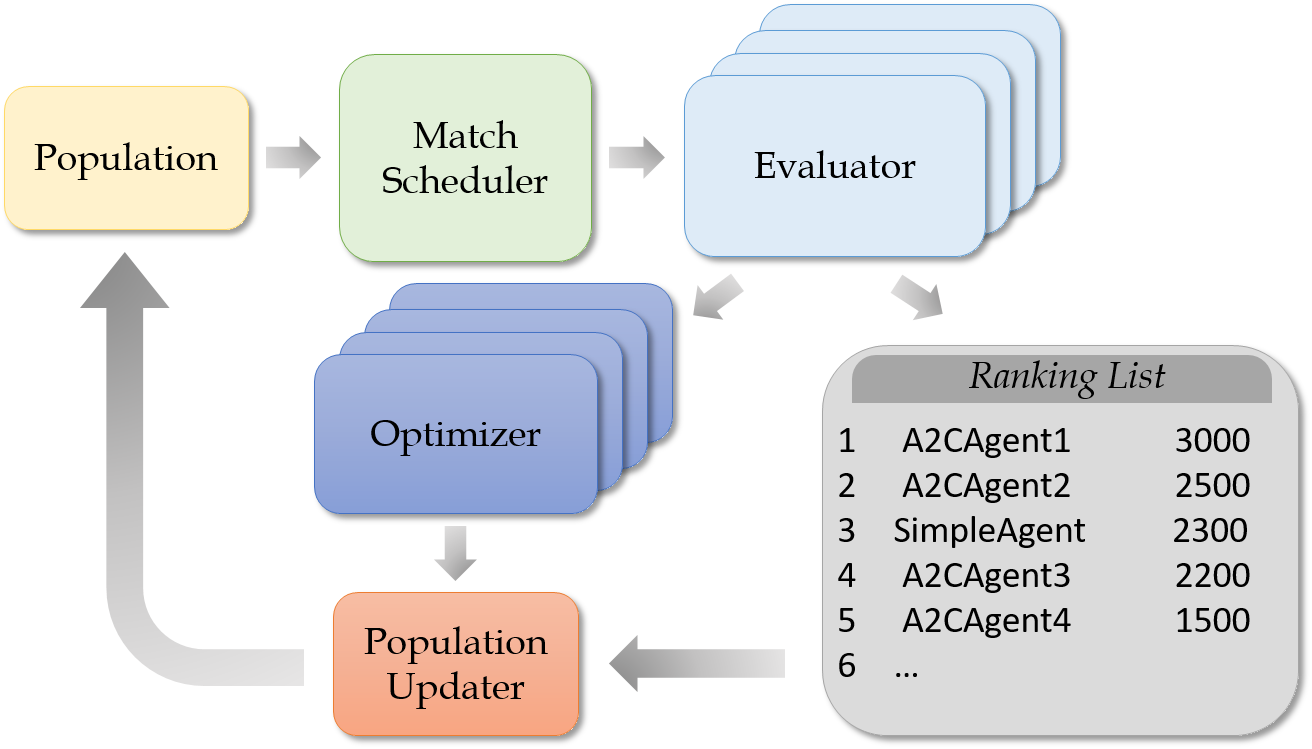}
  \centering
  \caption{The COMBAT infrastructure}
  \label{fig:combat}
\end{figure}

\textbf{Training Infrastructure}~~~
We first present the infrastructure of COMBAT in Figure~\ref{fig:combat}.
COMBAT consists of six modules.
The population module stores the indexes and the parameters of a population of agents.
Given the agent population, the match scheduler uses a round-robin method to generate a series of matches between any two agents in the population.
Then, the evaluator module deploys multiple workers in parallel to run the matches generated by the match scheduler, i.e., launching the same number of battles between those designated agents in the game environment accordingly.
The matches run on the evaluators are collected asynchronously: the trajectories of these matches are gathered and trained by the optimizer, while the ranking list assembles the results of these matches.
Next, the population updater is used to update the indexes of the agent population according to the results on the updated ranking list (i.e., the \emph{removable} function in the pseudo-code), and update the parameters of the agents in the population according to the outputs of the optimizers.
Lastly, the agent population module is synchronized based on the population updater.

Algorithm~\ref{alg:combat} gives the pseudo-code for more details.
The \emph{trainable} function is used to check whether an agent is trainable since some agents may be rule-based non-trainable.
The \emph{removable} function is used to check whether an agent should be removed from the population.
$\alpha$ is a hyperparameter to anneal the discount factor.
We assume that a maximum number of $T$ pickups is chosen for each worker.
Also, we omit the computation for $p_j, p_k, p_l$ in the pseudo-code, which is exactly the same as $p_i$.

COMBAT is also suitable for multiple people to work collaboratively, where each participant trains its own agent and compete with both the other trainable agents and the rule-based strong agents. With the ranking list telling which agents are performing better, we know which one's setting is working well. During the Pommerman competition, we exactly followed this way to communicate with each other, finding out the right way to tuning hyperparameters or trying different reward functions.

\begin{algorithm}[H]
\caption{Continual Match Based Training (COMBAT)}\label{alg:combat}
\begin{algorithmic}[1]
\Procedure{Train}{$\mathcal{P}$}\Comment{Initial population $\mathcal{P}$}
\State{initialize $\mathcal{R}, \alpha, \{\gamma_i\}_{p_i \in \mathcal{P}}$}
\For{t:=1 \text{to} T \text{(asynchronously in parallel)}}
    \State{Pick four agents $p_i, p_j, p_k, p_l$ from $\mathcal{P}$}
    \State{Play a match between $p_i,p_j, p_k, p_l$}
    \State{Update $\mathcal{R}$}
    \If{\emph{trainable}($p_i$)}
        \State{Update $p_i$}
    \EndIf
    \If{$p_i$ is converged}
        \State{$\gamma_i \gets \gamma_i + (1 - \gamma_i)*\alpha$}
    \ElsIf{\emph{removable}($p_i, \mathcal{R}$)}
        \State{Remove $p_i$ from $\mathcal{P}$}
    \EndIf
\EndFor
\State{Return $\mathcal{P}, \mathcal{R}$} 
\EndProcedure
\end{algorithmic}
\end{algorithm}







\textbf{State space}~~~
The original state space of Pommerman consists of three 11x11 matrices, respectively representing the position of different objects, bomb blast strength, and bomb life. However, such information is not available and covered by fog outside the purview of the current agent. Besides that, there are several scalars indicating teammate, enemies, ammo, blast strength and kick. To fit such state information with a convolutional neural network model, We encode all the state information in a 11x11x11 matrix with each channel representing different objects and their states. To make the states more precise, we also fix the bomb blast strength and the remaining time of its explosion for each bomb when it can be affected by the other bombs nearby.

\textbf{Action space}~~~
The original action space of Pommerman consists of six discrete actions: \emph{Stop, Up, Down, Left, Right and Lay a bomb}. One limitation of this kind of action space is that a local optimum arises where the agent avoids exploding itself by learning never to use the bomb action. Instead, the action space of our model is the board position plus bomb action, where the board position indicate the destination the agent predicts and bomb action indicates whether to lay a bomb at the current position. Therefore, we define an action space with 122 dimensions, where the first 121 dimensions are the flattened board positions, and the last one is the bomb action. Since the predicted positions are usually out of the reach of one step, Dijkstra algorithm is used to find the path to the destination.

\begin{figure*}
  \includegraphics[width=0.9\textwidth]{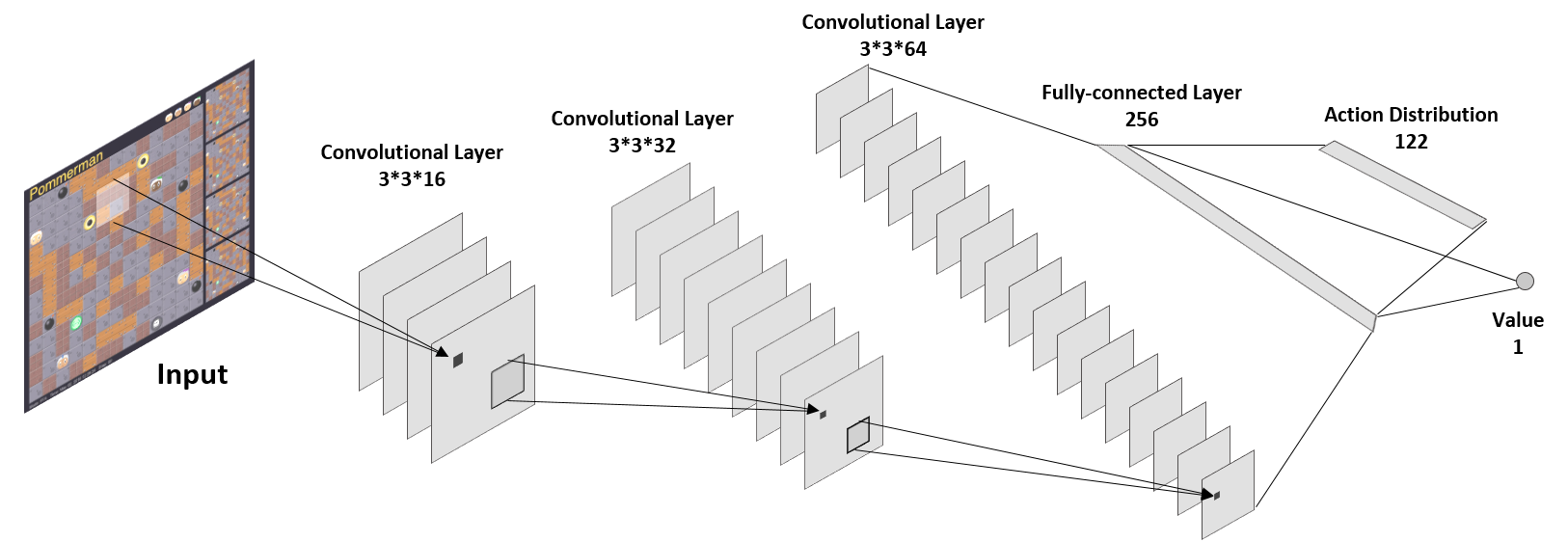}
  \centering
  \caption{The neural network structure}
  \label{fig:nn}
\end{figure*}

\textbf{Network Structure}~~~
For each agent, the neural network structure is illustrated in Figure~\ref{fig:nn}. After generating the square state space through the preprocessing step from the raw Pommerman frames, it is used as the input of the neural network. The first three hidden layers convolves 16, 32, and 64 3x3 filters with stride 1 of the input, all of which apply a rectifier nonlinearity~\cite{jarrett2009best, nair2010rectified}, respectively.
Then, the output of the third hidden layer is flattened, and it is connected to a fully-connected linear layer with the \emph{hyperbolic tangent} as the activation function.
The final outputs to either the action distribution or the value are connected with a fully-connected layer with their corresponding output sizes.

\textbf{Optimization Algorithm}~~~
We use A2C as the optimization algorithm for the Pommerman agent. Denote by the TD-loss $\delta = R(s,a) + \gamma \cdot V_{\phi}(s') - V_{\phi}(s)$.
The gradient of the actor network and the critic network are written as:
\begin{equation}
    \begin{aligned}
        \nabla_\theta J_\theta(\pi_\theta)
        & = \frac{1}{M}\sum_{s,a,s'\in D} \delta\cdot\frac{\partial\log\pi_\theta(a|s)}{\partial \theta}
    \end{aligned}
\end{equation}
\begin{equation}
    \begin{aligned}
        \nabla_\phi J_\phi
        &= \frac{1}{M}\sum_{s,a,s'\in D}\delta\cdot\frac{\partial V_\phi(s)}{\partial\phi}
    \end{aligned}
\end{equation}
where $s, a ,s'$ is a sequential state-action sample collected online, and $M$ is the size of a mini-batch. 

Following the spirit of continual learning, we adopt a policy-based algorithm advantage-actor-critic (A2C) as the basic infrastructure.
The reason why A2C is a good fit for continual learning is on two-fold.
First, the agent's policy represented by the actor network is independent of either the reward function or the hyperparameters used in the training phase. It can be used as the basis of generating meaningful behavior data, while we do not need to update its network structure when the reward function or the hyperparameters change.
Second, since the output of the critic network acts as the baseline when performing the policy update, the inaccuracy of estimating the critic network in A2C does not introduce any bias but related to the degree of variance reduction.
Theoretically, the variance is maximally reduced when the critic network approximates the expected cumulative reward. Therefore, when the reward function changes, the critic network should be updated to approximate the updated expected cumulative reward. Empirically, it does not bring significant effects to the learning results.

\section{Related work}
\label{sec:related}


Pommerman~\cite{resnick2018pommerman} is a multi-agent environment, which is stylistically similar to Bomberman. There are four agents traversing a grid world, and their goals are to have their team be the last remaining. They can move and lay bombs which, upon expiration, destroy any destructible objects in their ranges. Adversarial and cooperative elements are both encouraged in this environment. The Free-For-All (FFA) variant has at most one winner out of four agents, which fosters research on handling the non-equivalent Nash payoffs. The team variant encourages research on cooperation between two teammates with and without explicit communication channels. Based on Pommerman, learning-based methods and search-based methods have been proposed to tackle the problems. 

Backplay~\cite{resnick2018backplay} is proposed to increase the sample efficiency of RL in sparse reward settings. With a single demonstration, the curriculum for a given task can be constructed. The agent learns from the states at the end of each demonstration. During the training process, the starting point is being moved backward until the initial state is reached. Their experiment in Pommerman shows that Backplay provides significant gains in sample complexity with a stark advantage in sparse reward setting. Hybrid Search Agent~\cite{zhou2018hybrid} focused on search-based methods in Pommerman with resource-intensive forward models.
Their result shows that heuristic agent using depth-limited tree search can slightly outperform hand-made heuristics.
The depth-limited tree search agent with exploration-driven node selection can play significantly better than the provided \emph{SimpleAgent}.

Continual learning~\cite{parisi2018continual}, also known as lifelong learning, with neural network has been studied for quite a long time in the literature. The idea of continual learning is to learn a neural network to complete multiple tasks continually. The major difficulty in continual learning is the phenomenon of catastrophic forgetting when learning on new tasks. Existing work either define novel neural network structures or propose a more general regularizer in the loss function to maintain the elasticity of agents. However, when the experience exposed to the agent are correlated, the events of catastrophic forgetting are not exhibited frequently during the time of learning on new tasks. In Pommerman, suppose that we define each task as a particular skill that an agent is expected to learn such as the task of learning to pick up a power-ups, or the task of learning to kick a bomb. Then, these tasks are highly correlated and may appeared together in a single battle. As a result, pommerman itself is a natural environment for lifelong learning without suffering from catastrophic forgetting.

Hyperpamater search is a mature topic in the literature, where bayesian optimization and meta-learning are the most dominated methods.
Next, we highlight some of the novel techniques published in recent years below.
Population based training (PBT) ~\cite{jaderberg2017population} can be regarded as a type of meta learning~\cite{lemke2015metalearning}, where humans do not interfere with the learning process in general. The hyperparameter search in PBT requires no sequential runs, and used few computational resources than traditional grid search or sequential optimization methods.
However, while PBT stabilizes the training process based on the complete set of hyperparameters,
COMBAT concerns more about the hyperparameters that are essential for continual or lifelong learning.
The discount factor, considered in Pommerman, is a typical example of this kind: different settings may lead to completely different results of convergence.
Self-play learning~\cite{silver2017mastering} is a popular technique of making the competitive environment always stays in a suitable level of difficulty for training agents. While the self-play framework provides a natural curriculum for training agents, it also leads to a more non-stationary environment, where the goal of optimization is changing all the time along the training process. OpenAI proposed to choose opponents randomly from the historically saved models to stablize the training process~\cite{bansal2017emergent}. Likely, COMBAT consider a similar strategy, simultaneously training multiple agents and allow them to compete with each other.
Meta-gradient reinforcement learning~\cite{xu2018meta} considered a gradient-based method to optimize the RL-related hyperparameters such as the discount factor and the number of steps to bootstrap. In our current implementation of COMBAT, we only consider the discount factor in our hyperparameter optimization, while in the future work we may consider the other RL-related hyperparameters.

\section{Experiments: COMBAT in Pommerman}
\label{sec:exp}

We apply A2C with COMBAT to perform the maximization of cumulative reward in the Team environment.
Although the optimization of a particular POMDP is sensitive to hyperparameters, including learning rate, minibatch size, discount factor, and coefficients of policy loss, value loss, and entropy regularizer, COMBAT makes the final results more stable in a long-term training process.

\subsection{Experimental Setup}
\label{subsec:expsetup}

\textbf{Hyperparameters}~~~
COMBAT is only used to fine-tune the discount factor $\gamma$ in learning our Pommerman agent. The initial discount factor is $0.5$ and it will be increased gradually during the learning process.
We used the same network structure as Figure~\ref{fig:nn}.
For other hyperparameters, the coefficient of value loss is set to $0.5$ and the coefficient of entropy regularizer is set to $0.01$, which is then linearly decayed to $0.05$. The learning rate is $0.0005$ during the whole learning process without any annealing schedule.
The minibatch of each agent is collected every $256$ timesteps. An agent is updated by gradient descent using Adam optimizer~\cite{kingma2014adam} per $10$ seconds, where the gradient update formula is provided by A2C.

\textbf{Match Schedule}~~~
The match scheduler is fundamental in COMBAT.
The population of agents comprises two kinds of agents, the trainable agents, and the rule-based agents.
Each trainable agent is a deep neural network, while the rule-based agents include \emph{SimpleAgent} and \emph{SuperAgent}.
Rule-based agents play an essential role of out training: it is not only a good opponent for the trainable agents but also a reference in the ranking list for the trainable agents.
\emph{SimpleAgent}, an officially provided rule-based agent, is initially placed in the population.
\emph{SuperAgent}, an augmented version of \emph{SimpleAgent} implemented by ourselves, is added to the population in the later training phase.
We have a population of $9$ agents when participating in the competition, where there are $8$ trainable agents and $1$ rule-based agent.
A round-robin method is used to generate the competition schedule, but it is not a pure round-robin since we gave the rule-based agent a much higher probability to be chosen as one of the participants.

\begin{figure}
  \includegraphics[width=0.48\textwidth]{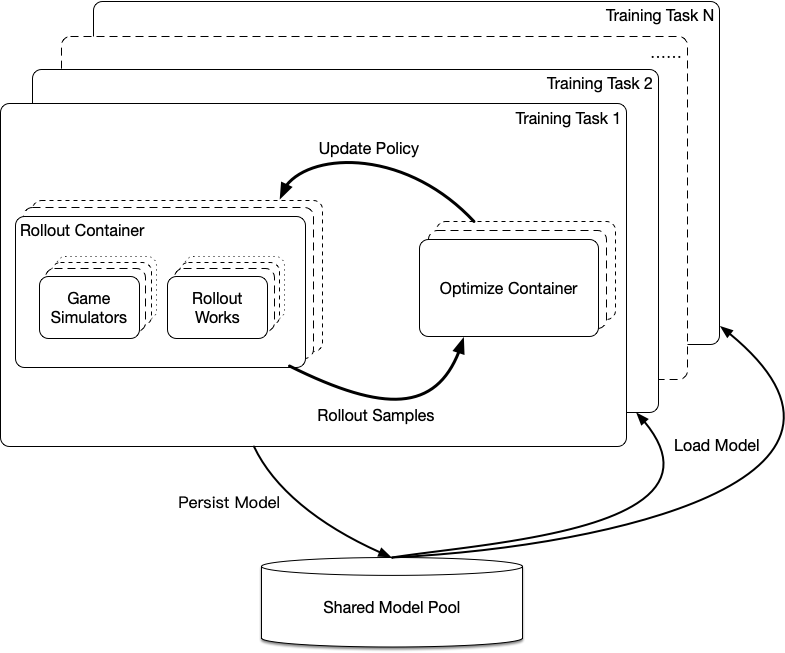}
  \centering
  \caption{The framework of our distributed asynchronous parallel system}
  \label{fig:system}
\end{figure}

\textbf{Distributed Asynchronous Parallel System}~~~
To run the COMBAT framework, we used 220 CPU cores and 32 GPU cores. Our asynchronous distributed parallel system is explained in Figure~\ref{fig:system}.
From the systematic perspective, we divide our system into two parts: the rollout generators and the policy optimizers. The rollout generators run many game simulators in parallelism, generating samples that are transmitted to the policy optimizer. The policy optimizer is responsible for computing the gradient and updating the parameters of the agents. Both of them have been built as Docker containers and deployed by the Kubernetes~\cite{bernstein2014containers}. With this methods, we reduce the cost of paralleling multiple tasks across the computational resources.

Moreover, since the agents need to be stored and shared persistently, the Ceph~\cite{weil2006ceph} was chosen as our distributed block storage solution, which was configured as a shared model pool. All the checkpoints generated by the training tasks can be stored in this pool and loaded by tasks on the other processes to initialize the training tasks. The system will relaunch when the error encountered, and restore all the tasks with the latest checkpoint.

\subsection{Experimental Results}

\begin{figure*}[t!]
  \includegraphics[width=0.8\textwidth]{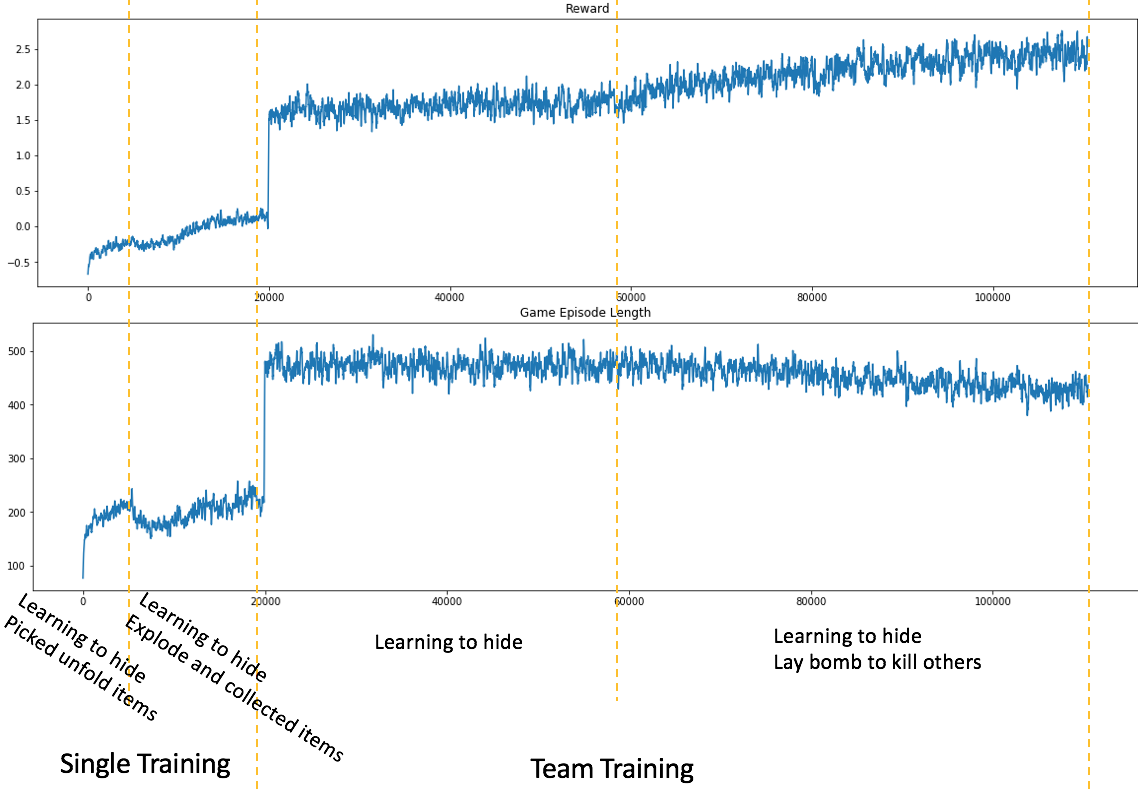}
  \centering
  \caption{The experimental results of training the \emph{Navocado} agent}
  \label{fig:navocado}
\end{figure*}

Figure~\ref{fig:navocado} presents the learning curve of our agent \emph{Navocado} which won the top 1 learning agent in the NeurIPS 2018 Pommerman competition.
The upper sub-figure in Figure~\ref{fig:navocado} shows the curve of reward, and the lower one in Figure~\ref{fig:navocado} shows the curve of Game Episode length.
The x-axis of both sub-figures is referred to the number of training iterations.
The learning process can be divided into four stages.
In the first stage, only one single agent was trained, and we used the default reward function. During this period, the agent learned to hide from the bomb explosions, while picking up power-ups that have been unfolded.
In the second stage, we added an extra reward of picking power-ups.
During this period, the agent learned to explode wooden walls actively and collected power-ups more efficiently.
Notice that in both stages, we set a \emph{SimpleAgent} as its teammate.
Then, in the third stage, we started the team training process where the \emph{SimpleAgent} teammate is replaced by a trainable agent such that the agents at the same team are trained collectively, but we did not allow the agent to kick the bomb at this stage.
During this period, it takes a much longer time to finish a match, and the agent learned a better skill of escaping from the bomb explosion.
Lastly, in the fourth stage, we allow both agents in the team to kick the bomb.
During this period, we found that the agents learned to attack more progressively. Notice that during this period, the Game Episode Length gradually decreased dramatically from 500 to 400.




\subsection{Examples}
\begin{figure*}
\includegraphics[width=0.8\textwidth]{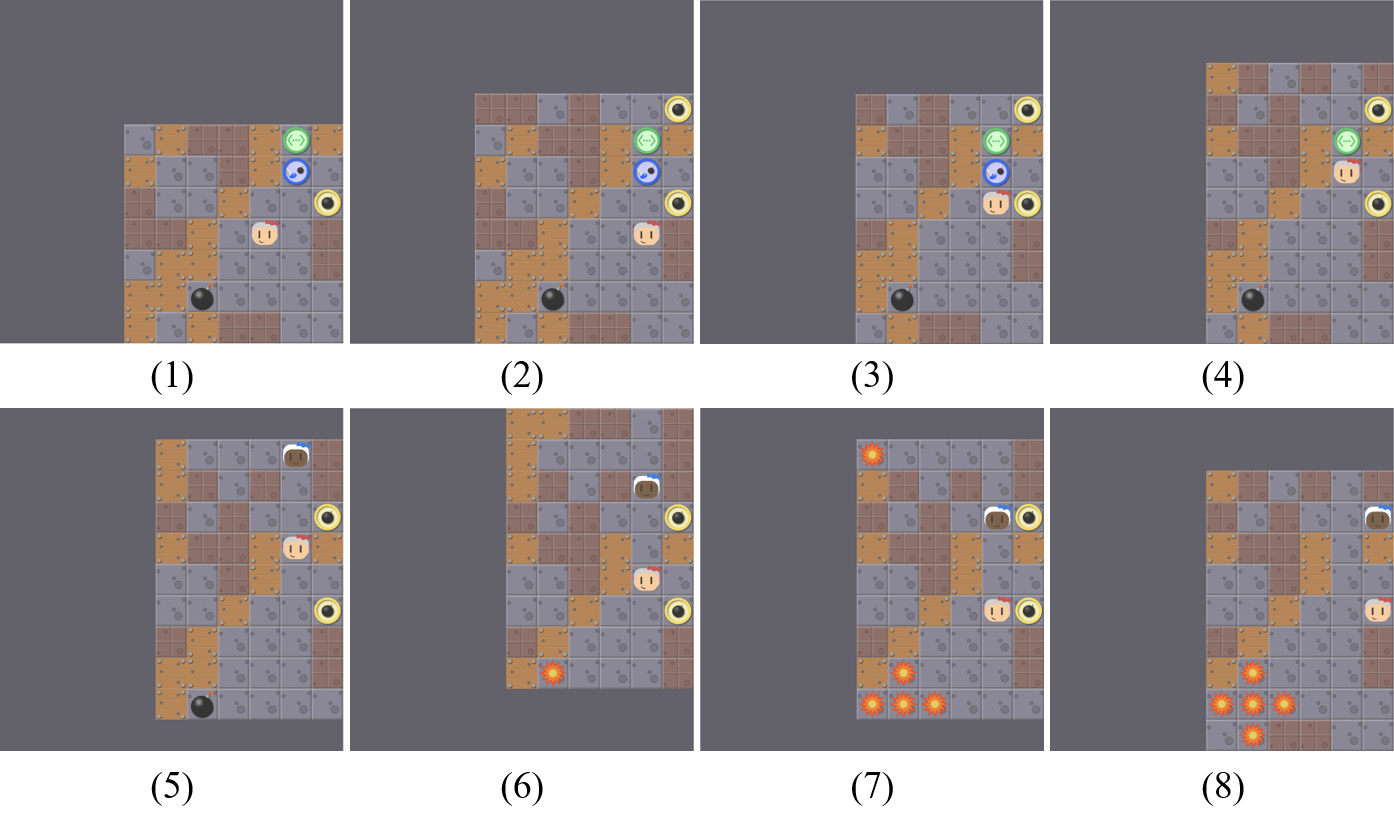}
\centering
\caption{The Pickup Movement: the agent can navigate with the shortest path to pick up various power-ups.}
\label{fig:pickup}
\end{figure*}

\begin{figure*}
\includegraphics[width=0.8\textwidth]{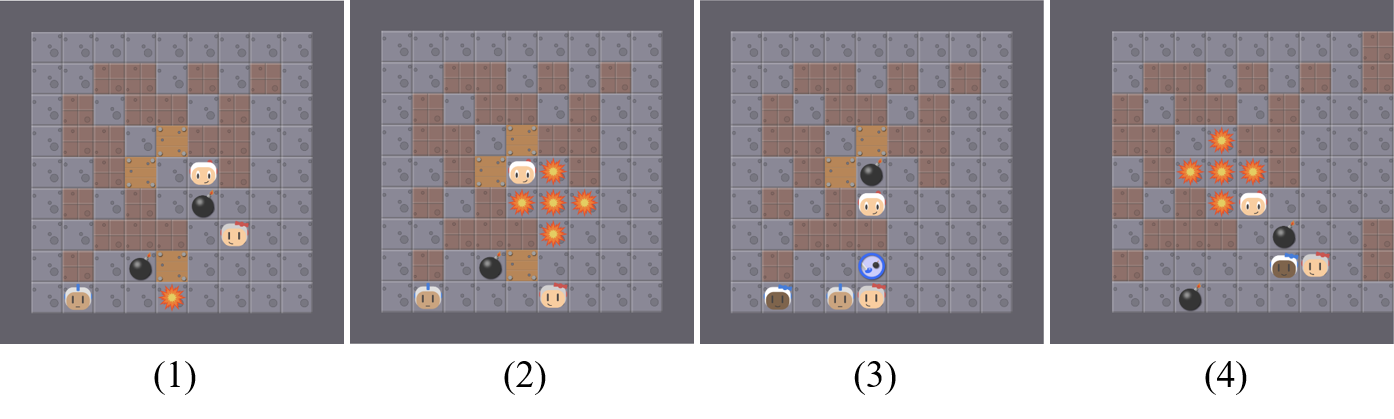}
\centering
\caption{The Hide-from-bomb Movement: the agent stays close to the bomb and hides from the bomb explosion.}
\label{fig:hide}
\end{figure*}

\begin{figure*}
\includegraphics[width=0.8\textwidth]{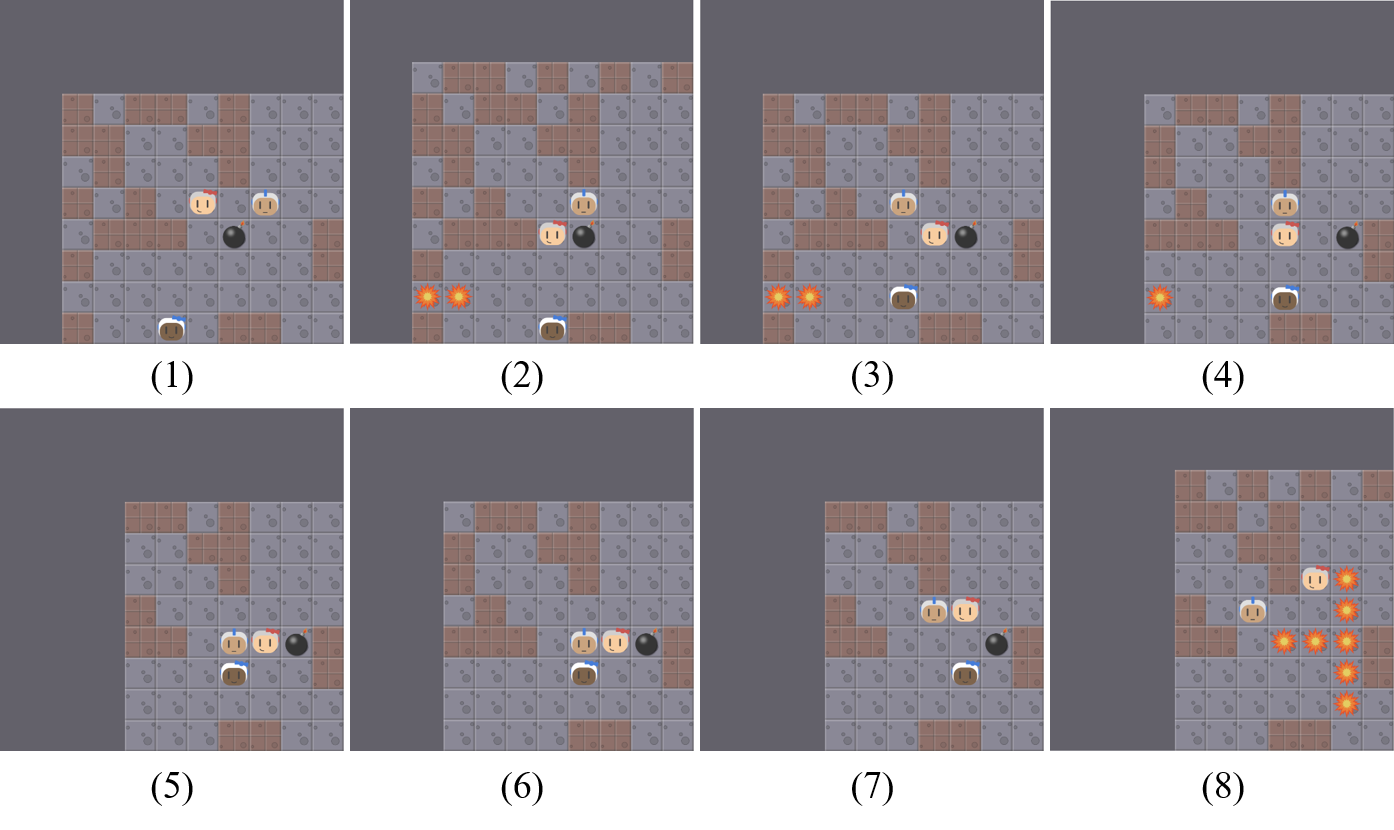}
\centering
\caption{The kick-bomb Movement: the agent can kick the bomb to the direction where the opponent locates.}
\label{fig:kick}
\end{figure*}

In the following, we summarize some skills of the agent we observed when the agent is playing against other agents.
In Figure~\ref{fig:pickup}, the agent reveals its ability to pick up a number of power-ups actively. We can see that the agent was taking the shortest path to pick up the items. Also, when the agent found that an enemy is close to an item, it chose to move backward and pick the item that is far away from the opponent agent.
In Figure~\ref{fig:hide}, we show a scenario that the agent can hide from the bomb explosion. Note that the agent stayed close to the bomb until the bomb explodes.
In Figure~\ref{fig:kick}, the agent learns to kick the bomb laid by the opponent to the direction where the opponent locates. By staying close to the bomb, the agent makes the opponent, which attempted to lay a bomb close to it, more likely to be destroyed by the bomb explosion.

\section{Discussion}
\label{sec:discuss}

We also observed some interesting results during the process of training our agents.
Firstly, the agents learned the skill of escaping from a bomb explosion at the beginning of the learning process, but it becomes much slower for them to learn other skills.
The early period of fast learning is similar to the developmental theory discovered in human learning~\cite{lenneberg2014foundations}, where infants have a critical period of learning fast from their experiences.
Secondly, the agents learned to kick the bomb laid by the opponent more actively than its own bomb. Since the bomb explosion is indistinguishable, the action of ``lay a bomb'' is not only the most effective way for an agent to attack actively but also a dangerous behavior that may kill the agent itself. Thus, a smart agent should lay a bomb in the right place. From this perspective, our agent may intend to ``place'' the opponent's bomb in the right place by kicking the bomb.
Thirdly, we found that if the agent did not learn the basic skill such as laying a bomb near the wooden wall in the early stage, when the discount factor is tuned, the agent can still learn the advanced skills such as kicking a bomb or tempting the opponent to killed by its bomb. However, the basic skill just cannot be learned anymore even if we tuned the discount factor back to the initial value at the later stage. This result may indicate that a complete set of basic skills should be learned before we tuned the hyperparameters for the advanced skills which are related to the long-term planning.


\section{Conclusions}
\label{sec:concl}
In the paper, we present a COMBAT framework for training Pommerman agents for participating the NeurIPS 2018 Pommerman Competition.
We use A2C for updating the deep neural network of the Pommerman agents, while a we design a match scheduling method for better training strong agents in a long-term learning process.
Empirically, the agent kept on improving its performance as long as dozens of days and won the top 1 learning agent in the Pommerman Competition.

There is still a huge space for our agent to improve in the future. Without considering the feature of partial observability, it is hard for our agent to learn how to collaborate with its teammate when its teammate is out of its sight. The basic idea is to model the problem as a multi-agent POMDP rather than an MDP.

\bibliography{mybib}
\bibliographystyle{plain}

\end{document}